%% file: ood.tex
\documentclass{article} 
\usepackage{iclr2021_conference,times}

\input{math_commands.tex}

\usepackage{hyperref}
\usepackage{url}

\usepackage{pgfgantt}
\usepackage{subfigure}
\usepackage{gensymb}
\usepackage{amsmath}
\usepackage{adjustbox} 
\usepackage{amsfonts}

\usepackage{multirow}
\usepackage{todonotes}

\usepackage{array}
\usepackage{graphicx}
\usepackage{multirow}
\usepackage{wrapfig}

\usepackage{tikz}
\usetikzlibrary{matrix,shapes,arrows,positioning,chains,arrows.meta,calc,decorations.markings,math}
\tikzstyle{process} = [rectangle, minimum width=3cm, minimum height=1cm, text centered, draw=black]
\tikzstyle{arrow} = [thick,->,>=stealth]
\tikzstyle{io} = [trapezium, trapezium left angle=70, trapezium right angle=110, text centered]

\newcommand{\DPNminus}{DPN$^-$ }
\newcommand{\D}{\mathbb{D}}

\title{Out-of-distribution detection in satellite image classification}
\author{Jakob Gawlikowski  \\
	Institute of Data Science\\
	German Aerospace Center, Jena, Germany\\
	\texttt{jakob.gawlikowski@dlr.de} \\
	\And
	Sudipan Saha \\
	Data Science in Earth Observation \\
    Technical University of Munich, Germany \\
	\texttt{sudipan.saha@tum.de} \\
	\AND
	Anna Kruspe \\
	Data Science in Earth Observation \\
    Technical University of Munich, Germany \\
	\texttt{anna.kruspe@tum.de} \\
	\And
	Xiao Xiang Zhu \\
	Remote Sensing Technology Institute \\
	German Aerospace Center, We{\ss}ling, Germany \\
	\texttt{xiaoxiang.zhu@dlr.de}
}

\iclrfinalcopy
\begin{document}
%
\maketitle
\begin{abstract}
In satellite image analysis, distributional mismatch between the training and test data may arise due to several reasons, including
unseen classes in the test data and differences in the geographic area. Deep learning based models may behave in unexpected 
manner when subjected to test data that has such distributional shifts from the training data, also called out-of-distribution (OOD) examples. Predictive 
uncertainly analysis is an emerging research topic
which has not been explored much in context of satellite image analysis.  Towards this, we adopt a  Dirichlet Prior Network based  model to quantify distributional uncertainty of deep learning models for remote sensing. The approach seeks to maximize the representation gap between the in-domain and OOD examples for a better identification
of unknown examples at test time. Experimental results on three exemplary test scenarios show the efficacy of the model  in satellite image analysis.
\end{abstract}

\section{Introduction}
\label{sectionIntroduction}

Deep learning has revolutionized the field of remote sensing in the last few years \cite{ball2017comprehensive, mou2021deep, saha2019unsupervised}. Most of the
satellite image analysis approaches assume that test data is similarly distributed as the training data on which the model is trained. However, this assumption rarely holds in practice. Remote sensing
deals with a large number of acquisition sensors  operating across a variety of different geographies. Moreover, some landscape classes seen be seen in only some geographic areas.   Deep learning models
are likely to fail or behave in an unexpected way when faced with open-set classes. A deep model trained on images from agricultural area will likely fail when asked to predict urban images comprising unseen classes. Similarly, deep models behave in unexpected way when fed with data from seen classes but with considerable geographic variation. For example, 
European and Asian urban areas exhibit significantly different semantics and a model trained on one may likely fail on the another, forcing to use geography-wise different models \cite{saha2020change}. When deep learning based systems fail, they do not provide sufficient cue to the user and can 
give a wrong prediction, yet with high confidence. To address this issue, predictive uncertainty estimation has recently emerged as a research topic in the machine learning community \cite{malinin2018predictive}. Uncertainty
estimation informs users about the confidence on a prediction, thus gives a hint on the reliability of such systems and possible weaknesses.
\par
Deep learning based classification models are prone to predictive uncertainties from three different sources \cite{malinin2018predictive}: model or epistemic uncertainty, data or aleatoric uncertainty, and distributional uncertainty.  In remote sensing distributional uncertainty may arise due to various reasons, as unseen classes, geographic
differences, and sensor differences. Considering its high relevance in satellite image analysis, our work focuses on distributional uncertainty \cite{gal2016uncertainty}. 
\par
Our work is based on a Dirichlet Prior Network (DPN) that separately models different aforementioned uncertainty types. The Dirichlet distribution is a distribution over the categorical distribution, i.e. it can model uncertainty on a soft-max output of a classification model. DPNs separate in-distribution and OOD examples by producing sharp Dirichlet distributions for in-domain examples (low deviation in the softmax output) while producing flat Dirichlet distributions for OOD ones (high deviation in the softmax output) \cite{malinin2018predictive}. In particular, we base our work on an extension of the DPN classifier \cite{nandy2020towards} that focuses on increasing the representation gap between in-domain and OOD examples. 
We experimentally show that the proposed approach is able to detect OOD examples in remote sensing images, thus improving the reliability and robustness of deep learning based models in remote sensing. To the best of our knowledge this is the first work that specifically addresses out-of-distribution detection in remote sensing.

\begin{wrapfigure}{r}{5.0cm}
	\centering
	\subfigure[]{%
		\includegraphics[height=1.4 cm]{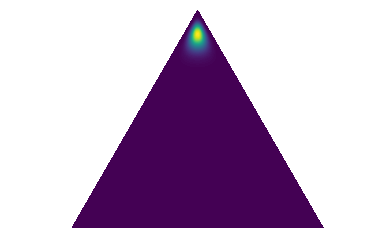}
		\label{simplexConfident}
	}%
	\subfigure[]{%
		\includegraphics[height=1.4 cm]{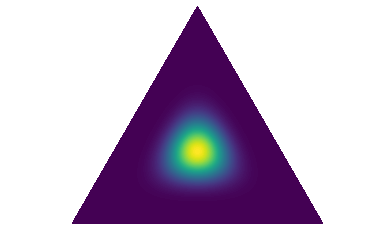}
		\label{simplexDataUncertaintyIdeal}
	}%

	\subfigure[]{%
		\includegraphics[height=1.4 cm]{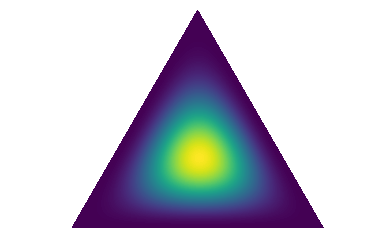}
		\label{simplexOODDpn}
	}%
	\subfigure[]{%
		\includegraphics[height=1.4 cm]{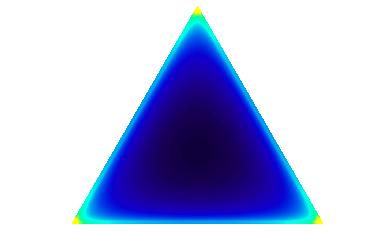}
		\label{simplexOODAfterPrecisionRegularization}
	}
	\caption{Different desired predictive uncertainties shown over the simplex (cf. \cite{nandy2020towards}): (a) In-domain confident, (b) In-domain aleatoric uncertainty, (c) OOD (with DPN \cite{malinin2018predictive}), (d) OOD (with \DPNminus \cite{nandy2020towards}).}
	\label{figureSimplexRepresentation}
\end{wrapfigure}

\section{DPN for satellite image analysis}
\label{sectionProposedMethod}
In satellite image classification, images $x$ and their corresponding labels $y$ can be characterized using their distribution $p(x,y)$. In practice, we only have a finite data set $\D = \{x_j,y_j\}_{j=1}^N$ corresponding to the distribution $p(x,y)$. Since the training data is a random subset and the training process is also affected by randomness, Bayesian neural networks model the parameters $\theta$ of a neural network as a random variable. For a classifier with parameters $\theta$ the predictive uncertainty on a prediction $\omega$ is then given by $p(y=\omega|x^{*},\D)=p(y=\omega|x^*,\theta)p(\theta|\D)$. \\
The sources of predictive uncertainty \cite{malinin2018predictive} can be broadly categorized into the following three categories: epistemic or model uncertainty, aleatoric or data uncertainty, and distributional uncertainty.
Distributional uncertainty is likely in remote sensing due to differences caused by new classes in the target data, geographic shift, and multi-sensor differences. 
Approaches as Bayesian Neural Networks and deep ensembles consider the distributional uncertainty as part of the epistemic uncertainty. These approaches seek to explicit predict the aleatoric uncertainty and to quantify the epistemic uncertainty by performing several predictions with different model parameters \cite{lakshminarayanan2017simple}.
\par
Dirichlet distributions are popularly used as a prior distribution in Bayesian learning. Malinin and Gales \cite{malinin2018predictive} proposed Dirichlet Prior Networks (DPN) that efficiently mimic the behavior of Bayesian networks by parameterizing a Dirichlet distribution over the categorical distribution given by a soft-max classification output. Convenient to remote sensing applications, any neural network with soft-max activation can be considered as a DPN. A Dirichlet distribution over $K$ classes is characterized by concentration parameters $\{\alpha_1,...,\alpha_K\}>0$. For a DPN the concentration is given by the exponentials of the network's logit values $z$,
\begin{equation}
	\alpha_k = exp(z_k(x^{*}))~.
\end{equation}
The sum of the concentrations $\alpha_0 = \alpha_1+...+\alpha_K$ is called the precision of the distribution. The larger the precision, the sharper is the Dirichlet distribution.

For in-domain samples where the classifier is confident, DPNs aim to produce
uni-modal distribution at the corner of the solution simplex with the correct class (Figure \ref{simplexConfident}) \cite{malinin2018predictive}. For in-domain samples with high data uncertainty DPNs aim to produce a sharp distribution at the center (Figure \ref{simplexDataUncertaintyIdeal}) and for OOD data a flat distribution (Figure \ref{simplexOODDpn}). 
However, for in-domain examples with high aleatoric uncertainty among multiple classes, DPNs could also produce flat Dirichlet distributions \cite{nandy2020towards}, what often leads to representations which are indistinguishable from OOD examples. To overcome this, Nandy et al. \cite{nandy2020towards} proposed the \DPNminus approach. \DPNminus aims at learning a sharp multi-modal distribution ($\alpha_0<<1$) instead of a flat uni-modal distribution for OOD examples. Additional, Nandy et al. chose DPN parameters in a way, that the loss simplifies to the cross-entropy plus a precision regularization term. 

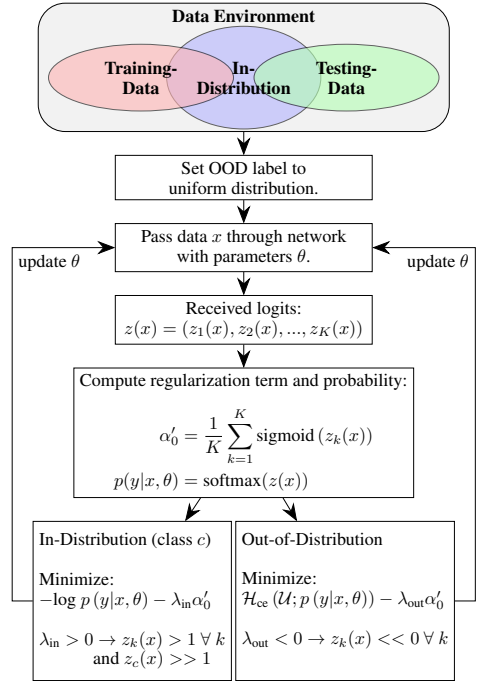
\begin{wrapfigure}{r}{5.0cm}
\label{fig:DNPminusDiagram}
	\centering
	\resizebox{0.45\textwidth}{!}{
	\begin{tikzpicture}
	   	\node [draw, rounded corners = 15pt, minimum width=8cm, minimum height=2.5cm, text depth=2cm, fill=gray!10,font=\bfseries] (data) at (4,5) {Data Environment};

 		\node [draw, ellipse, align=center, minimum width=3cm, minimum height=2cm, fill=blue!30, opacity=.6, text opacity=1,font=\bfseries](indata) at (4,4.8){};
	   	
	   	\node [draw, ellipse, align=center, minimum width=3.6cm, minimum height=1.3cm, fill=red!30,opacity=.6, text opacity=1,font=\bfseries](indata) at (2,4.8){ Training-\\ Data};
	   	
	   	\node [draw, ellipse, align=center, minimum width=3.6cm, minimum height=1.3cm, fill=green!30,opacity=.6, text opacity=1,font=\bfseries](indata) at (6,4.8){ Testing-\\ Data};
 		
 		\node [draw, ellipse, align=center, minimum width=3cm, minimum height=2cm, fill=blue!30, opacity=.0, text opacity=1,font=\bfseries](indata) at (4,4.8){ In-\\ Distribution}; 		
 		
 		\node [draw, below=0.45 of data, minimum width=5cm, align=center](1){Set OOD label to \\uniform distribution.};

 		\node [draw, below=.45 of 1, minimum width=5cm, align=center](2){Pass data $x$ through network \\with parameters $\theta$.};

 		\node [draw, below=0.45 of 2, minimum width=5cm, align=center](3){Received logits:\\ $z(x) = \left(z_1(x), z_2(x), ..., z_K(x)\right)$};

 		\node [draw, below=0.45 of 3, minimum width=5cm, align=center](4){Compute regularization term and probability:\\~\\
 		$\begin{aligned}
 		\alpha_0^\prime &= \frac{1}{K}\sum_{k=1}^K \text{sigmoid}\left(z_k(x)\right)\\
 		p(y|x,\theta)&=\text{softmax}(z(x))
 		\end{aligned}$};
 	
 		\draw[-{Stealth[scale=2.0]}] (data) edge (1) (1) edge (2) (2) edge (3) (3) edge (4);
 		
 		\node[draw, below = 1 of 4, minimum width=8.2cm, minimum height=2cm, opacity=0.0](dummy){};
 		\node[draw, right = 0.0 of dummy.west, minimum width=3.2cm, minimum height=3.1cm, align=left](in){In-Distribution (class $c$) \\~\\ 
 			
		Minimize: \\
		$-\text{log}~p\left(y\vert x, \theta\right)-\lambda_{\text{in}}\alpha_0^\prime$ \\~\\
		$\lambda_{\text{in}}>0 \rightarrow z_k(x)>1~\forall~k$ \\
 		~~~~~~~~~~~~\text{and }$z_c(x) >> 1$};
 	
 		\node[draw, left =  0.0 of dummy.east, minimum width=3.2cm, minimum height=3.1cm, align=left,text depth=0.5cm](out){Out-of-Distribution\\~\\ 
 			
		Minimize: \\
		$\mathcal{H}_{\text{ce}}\left({\mathcal{U}}; p \left(y\vert x, \theta \right)\right)-\lambda_{\text{out}}\alpha_0^\prime$ \\~\\
		$\lambda_{\text{out}}<0 \rightarrow z_k(x)<<0~\forall~k$};
 		
 		\draw[-{Stealth[scale=2.0]}] (4) edge (out);
 		\draw[-{Stealth[scale=2.0]}] (4) edge (in);
 		\draw[-{Stealth[scale=2.0]}] (in.west) -- +(-0.4,0) |- node [midway,below right]{update $\theta$}(2.west);
 		\draw[-{Stealth[scale=2.0]}] (out.east) -- +(0.4,0) |- node [midway,below left]{update $\theta$}(2.east);
	\end{tikzpicture}
}
\caption{\small{Visualization of the training procedure for the considered \DPNminus network.}}
\end{wrapfigure}

The precision regularization is achieved by introducing a bounded regularization term 
\begin{equation*}
\alpha_0^\prime = \frac{1}{K}\sum_{k=1}^K\text{sigmoid}(z_k({x}))
\end{equation*}
as a regularizer along with the cross-entropy loss. This gives the following two loss formulations for in-domain and OOD examples: 
\begin{equation}
\label{eqInDomainLoss}
\small
\mathcal{L}_{in}({\theta}; \lambda_{in}) := \mathbb{E}_{P_{in}(x, y)} \left[- \log p({y} | {x}, {\theta})
- \lambda_{in}\alpha'_0 \right]
\end{equation}
and
\begin{equation}
\label{eqOODDomainLoss}
\small
\mathcal{L}_{out}({\theta}; \lambda_{out}) :=\mathbb{E}_{P_{out}(x, y)} \left[\mathcal{H}_{ce} ( \mathcal{U}; p( {y} | {x}, {\theta}))
-  \lambda_{out}\alpha'_0 \right]~.
\end{equation}
$\mathcal{U}$ denotes the uniform distribution over all classes, $\mathcal{H}_{ce}$ denotes the cross-entropy function, and the precision is controlled by two hyper-parameters $\lambda_{in}>0$ and $\lambda_{out}<0$. The combined loss-function is given by
\begin{equation}
\label{eqTotalLoss}
\small
\mathcal{L}({\theta}; \gamma, \lambda_{in}, \lambda_{out}) = \mathcal{L}_{in}({\theta}, \lambda_{in}) 
+ \gamma \mathcal{L}_{out}({\theta}, \lambda_{out}),
\end{equation}
where in-domain and OOD samples are balanced by $\gamma>0$.
\par
For in-domain examples which are confidently predicted, the cross-entropy loss maximizes the logit value of the correct class. However, for in-domain samples with aleatoric uncertainty, the optimizer maximizes $\text{sigmoid}(z_k({x}))$ for all classes, thus yielding a flatter distribution. By choosing $\lambda_{out}<0$, \DPNminus produces uniform negative values for $z_k(x^*)$ for an OOD
example $x^*$. This leads to $\alpha_k << 1$ for all $k=1,...,K$, and thus an OOD sample yields a sharp multi-modal Dirichlet distribution with uniform weights at each corner of the simplex (Fig \ref{simplexOODAfterPrecisionRegularization}). Figures \ref{simplexDataUncertaintyIdeal} and  \ref{simplexOODAfterPrecisionRegularization} are more distinct over the simplex, making the OOD samples easier distinguishable from the in-domain ones. In Figure \ref{fig:DNPminusDiagram} a visualization of the training process of \DPNminus is given.
\section{Experiment and results}\label{sectionResult}

\input{input_files/ResultTable_big.tex}

\tikzset{
	mynode/.style={
		draw
		, align=left
		, execute at begin node=\setlength{\baselineskip}{10px}
		, anchor=south
	}
}
\begin{wrapfigure}{r}{8.0cm}

\label{fig:examples}

\begin{tikzpicture}[every node/.style={inner sep=2,outer sep=0.5}]

\node[inner sep=0pt] (C) at (0,0)
{\includegraphics[width=.07\textwidth]{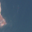}};
\node[below] at (C.south) {\footnotesize{Class G}};	
\node[right, align=left] at (C.east) {\footnotesize{MP: 0.105} \\ \footnotesize{MI: 2.118} \\ 
	\footnotesize{$\alpha_0$: 0.138}};

\node[inner sep=0pt] (D) at (3,0)
{\includegraphics[width=.07\textwidth]{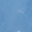}};
\node[below] at (D.south) {\footnotesize{Class G}};
\node[right, align=left] at (D.east) {\footnotesize{MP: 0.125} \\ \footnotesize{MI: 1.686} \\ 
	\footnotesize{$\alpha_0$: 0.602}};

\node[inner sep=0pt] (E) at (6,0)
{\includegraphics[width=.07\textwidth]{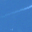}};
\node[below] at (E.south) {\footnotesize{Class G}};
\node[right, align=left] at (E.east) {\footnotesize{MP: 0.333} \\ \footnotesize{MI: 0.272} \\ 
	\footnotesize{$\alpha_0$: 11.43}};

\node[inner sep=0pt] (07) at (0,2)
{\includegraphics[width=.07\textwidth]{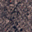}};
\node[below] at (07.south) {\footnotesize{Class 2}};	
\node[right, align=left] at (07.east) {\footnotesize{MP: 0.97} \\ \footnotesize{MI: $0.00002$} \\ 
	\footnotesize{$\alpha_0$: $\approx10^7$}};
	
\node[inner sep=0pt] (08) at (3,2)
{\includegraphics[width=.07\textwidth]{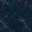}};
\node[below] at (08.south) {\footnotesize{Class 2}};
\node[right, align=left] at (08.east) {\footnotesize{MP: 0.77} \\ \footnotesize{MI: 0.0} \\ 
	\footnotesize{$\alpha_0$: $\approx10^{16}$}};
	
\node[inner sep=0pt] (09) at (6,2)
{\includegraphics[width=.07\textwidth]{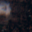}};
\node[below] at (09.south) {\footnotesize{Class 5}};
\node[right, align=left] at (09.east) {\footnotesize{MP: 0.246} \\ \footnotesize{MI: 1.537} \\ 
	\footnotesize{$\alpha_0$: 0.692}};


\end{tikzpicture}
\caption{Visualization of example samples from the left out 10\% of the training set of the So2Sat LCZ42 data set. The results are based on the \DPNminus network trained on urban (in-distribution) and vegetation (out-of-distribution) samples. One can clearly see the differences in the metrics. The two examples on the right side do not fit well into our assumptions, possibly caused by the clear edge in the water image and the blur in the urban image.}
\label{fig:LCZs}
\end{wrapfigure}

In order to evaluate the gap between in-domain and OOD samples we use the same measures as in \cite{nandy2020towards}, namely mutual information, maximum probability, and the precision $\alpha_0$. The general performance is characterized by \textit{area under the receiver operating characteristic} (AUROC) scores based on these three measures. 
\par
\textit{Test dataset:} We use the So2Sat LCZ42 dataset \cite{zhu2020so2sat} for evaluating the OOD detection performance. The dataset consists of local climate zone (LCZ) labels of approximately half a million Sentinel-2 
patches. Note that Sentinel-2 satellite images are significantly different from natural images (used in computer vision) having 13 spectral bands and 10 m/pixel spatial resolution.
The local climate zones are described by 17 classes, 1-10 corresponding to urban areas, A-F corresponding to non-urban areas, and G corresponding to water body.
We performed our experiments using following combinations:
\begin{enumerate}
\item Urban classes as in-domain data, non-urban ones as OOD data during training, and water body as OOD data during test.
\item Red channels (corresponding to all 17 classes) as in-domain, green channels as OOD during training, and blue channels as OOD during test.
\item Urban and vegetation classes as in-domain, rock and pavement as OOD during training, and water as OOD during test.
\end{enumerate}

\par
\textit{Deep architecture:} We used five sequential layers with 32, 64, 64, and 128 convolutional filters of size 3x3 each, followed by a dense layer of size 256. After each convolution layer, batch normalization is applied.  The networks are trained for 200 epochs.
\par
\textit{Comparison methods:} We consider a binary classifier trained to separate in-domain and OOD data. We evaluate the performance on a left-out 10\% subset of the training set (evaluation on seen regions) and on the  OOD samples from unseen regions. 
\par
\textit{Results:} In Table \ref{tab:results} the results based on 5 runs for each setting are presented and in Figure \ref{fig:examples} six examples are shown. The \DPNminus network clearly outperforms the binary classifier in separating in-domain and OOD examples on seen and unseen regions. The use of mutual information or the precision value contributes to increase the AUROC scores for the \DPNminus network for all test instances. Among the different considered cases, separating urban and vegetation classes is clearly most trivial, while the exclusion of single classes, as in test case 3, is significantly difficult. However, \DPNminus still perform satisfactorily for this task. 
\section{CONCLUSION}
\label{sectionConclusion}
In this paper, we quantified distributional uncertainty in deep learning models for satellite image analysis.  We tested the method on the So2Sat LCZ42 dataset considering open set classes and selected bands as OOD. Satellite images are 
significantly different from the natural images dealt in computer vision. It
is important to understand predictive uncertainty in context of satellite image analysis and our work is a first step towards it.

\bibliography{sigproc}
\bibliographystyle{iclr2021_conference}

\end{document}

%% file: math_commands.tex
\usepackage{amsmath,amsfonts,bm}

\def\eqref#1{equation~\ref{#1}}

\def\1{\bm{1}}

\DeclareMathAlphabet{\mathsfit}{\encodingdefault}{\sfdefault}{m}{sl}
\SetMathAlphabet{\mathsfit}{bold}{\encodingdefault}{\sfdefault}{bx}{n}



%% file: input_files/ResultTable_Big.tex
\begin{table*}[!ht]
	\centering
	\label{tab:results}
	\vspace{0.2cm}
	\begin{tabular}{p{0.9cm}p{1.7cm}p{1.7cm}p{1.7cm}p{1.7cm}p{1.7cm}}
		 & &\multicolumn{2}{c}{\DPNminus network}
		&\multicolumn{2}{c}{Binary Classifier}\vspace{0.2cm}\\
		 
		& &\multicolumn{1}{p{1.6cm}}{\centering Testing Data Set} 		&\multicolumn{1}{p{1.6cm}}{\centering Left out 10\% of Training Set} 	
		&\multicolumn{1}{p{1.6cm}}{\centering Testing Data Set} 		&\multicolumn{1}{p{1.6cm}}{\centering Left out 10\% of Training Set}\\ \hline
		\multirow{3}{1.0cm}{Test Case 1}
		&Max. Prob.		&$95.51\pm1.63$		&$98.66\pm0.37$		&$90.67\pm1.10$		&$91.87\pm1.76$		\\
		&Mutual Info 	&$96.28\pm0.57$		&$99.24\pm0.32$		&- 	&-				\\
		&$\alpha_0$		&$96.26\pm0.51$		&$99.23\pm0.33$		&-	&-			\\ \hline
	
		\multirow{3}{1.0cm}{Test Case 2}		
		&Max. Prob.		&$73.99\pm3.59$		&$87.88\pm2.54$		&$60.31\pm4.53$		&$73.79\pm4.58$		\\
		&Mutual Info 	&$81.81\pm1.68$		&$93.85\pm1.06$		&- 		&-				\\
		&$\alpha_0$		&$85.15\pm1.94$		&$95.01\pm0.88$		&-		&-			\\ \hline
		
		\multirow{3}{1.0cm}{Test Case 3}		
		&Max. Prob.		&$83.15\pm3.46$		&$92.27\pm2.88$		&$53.73\pm9.86$			&$86.42\pm4.93$		\\
		&Mutual Info 	&$87.03\pm1.21$		&$95.62\pm2.80$		&- 		&-				\\
		&$\alpha_0$		&$86.94\pm1.17$		&$95.53\pm2.75$		&-		&-			\\ 
	\end{tabular}
	\caption{AUROC scores of the \DPNminus and a binary classifier baseline network. The scores are based on maximum probability, mutual information, and precision for the \DPNminus. For the binary classifier, only the maximum probability is considered, since $\alpha_0$ is related to the Dirichlet distribution and mutual information can not be used for a binary variable. The results are given as mean and standard deviation of five runs.}
\end{table*}

%% file: ood.bbl
\begin{thebibliography}{9}
\providecommand{\natexlab}[1]{#1}
\providecommand{\url}[1]{\texttt{#1}}
\expandafter\ifx\csname urlstyle\endcsname\relax
  \providecommand{\doi}[1]{doi: #1}\else
  \providecommand{\doi}{doi: \begingroup \urlstyle{rm}\Url}\fi

\bibitem[Ball et~al.(2017)Ball, Anderson, and Chan]{ball2017comprehensive}
John~E Ball, Derek~T Anderson, and Chee~Seng Chan.
\newblock Comprehensive survey of deep learning in remote sensing: theories,
  tools, and challenges for the community.
\newblock \emph{Journal of Applied Remote Sensing}, 11\penalty0 (4):\penalty0
  042609, 2017.

\bibitem[Gal(2016)]{gal2016uncertainty}
Yarin Gal.
\newblock Uncertainty in deep learning.
\newblock \emph{University of Cambridge}, 1\penalty0 (3):\penalty0 4, 2016.

\bibitem[Lakshminarayanan et~al.(2017)Lakshminarayanan, Pritzel, and
  Blundell]{lakshminarayanan2017simple}
Balaji Lakshminarayanan, Alexander Pritzel, and Charles Blundell.
\newblock Simple and scalable predictive uncertainty estimation using deep
  ensembles.
\newblock \emph{Advances in neural information processing systems},
  30:\penalty0 6402--6413, 2017.

\bibitem[Malinin \& Gales(2018)Malinin and Gales]{malinin2018predictive}
Andrey Malinin and Mark Gales.
\newblock Predictive uncertainty estimation via prior networks.
\newblock \emph{arXiv preprint arXiv:1802.10501}, 2018.

\bibitem[Mou et~al.(2021)Mou, Saha, Hua, Bovolo, Bruzzone, and
  Zhu]{mou2021deep}
Lichao Mou, Sudipan Saha, Yuansheng Hua, Francesca Bovolo, Lorenzo Bruzzone,
  and Xiao~Xiang Zhu.
\newblock Deep reinforcement learning for band selection in hyperspectral image
  classification.
\newblock \emph{arXiv preprint arXiv:2103.08741}, 2021.

\bibitem[Nandy et~al.(2020)Nandy, Hsu, and Lee]{nandy2020towards}
Jay Nandy, Wynne Hsu, and Mong~Li Lee.
\newblock Towards maximizing the representation gap between in-domain \&
  out-of-distribution examples.
\newblock \emph{arXiv preprint arXiv:2010.10474}, 2020.

\bibitem[Saha et~al.(2019)Saha, Bovolo, and Bruzzone]{saha2019unsupervised}
Sudipan Saha, Francesca Bovolo, and Lorenzo Bruzzone.
\newblock Unsupervised deep change vector analysis for multiple-change
  detection in vhr images.
\newblock \emph{IEEE Transactions on Geoscience and Remote Sensing},
  57\penalty0 (6):\penalty0 3677--3693, 2019.

\bibitem[Saha et~al.(2020)Saha, Bovolo, and Bruzzone]{saha2020change}
Sudipan Saha, Francesca Bovolo, and Lorenzo Bruzzone.
\newblock Change detection in image time-series using unsupervised lstm.
\newblock \emph{IEEE Geoscience and Remote Sensing Letters}, 2020.

\bibitem[Zhu et~al.(2019)Zhu, Hu, Qiu, Shi, Kang, Mou, Bagheri, H{\"a}berle,
  Hua, Huang, et~al.]{zhu2020so2sat}
Xiao~Xiang Zhu, Jingliang Hu, Chunping Qiu, Yilei Shi, Jian Kang, Lichao Mou,
  Hossein Bagheri, Matthias H{\"a}berle, Yuansheng Hua, Rong Huang, et~al.
\newblock So2sat {LCZ}42: A benchmark dataset for global local climate zones
  classification.
\newblock \emph{arXiv preprint arXiv:1912.12171}, 2019.

\end{thebibliography}
